\documentclass{article}

\PassOptionsToPackage{numbers, compress}{natbib}
\usepackage[preprint]{neurips_2024}

\usepackage[utf8]{inputenc} %
\usepackage[T1]{fontenc}    %
\usepackage[colorlinks,bookmarks=false]{hyperref}
\usepackage{url}            %
\usepackage{booktabs}       %
\usepackage{amsfonts}       %
\usepackage{nicefrac}       %
\usepackage{microtype}      %
\usepackage{xcolor}         %

\usepackage{colortbl}
\usepackage{tikz}
\usepackage{booktabs}
\usepackage{multicol}
\usepackage{multirow}
\usepackage{color}
\usepackage{caption}
\usepackage{subcaption}
\usepackage{hhline}
\usepackage{pifont}
\usepackage{threeparttable}
\usepackage{makecell}
\usepackage{algorithm} 
\usepackage{listings}
\usepackage{amsfonts,amssymb}
\usepackage{amsmath}
\usepackage{graphicx}
\usepackage{lipsum}
\usepackage{arydshln}
\usepackage[accsupp]{axessibility}
\usepackage{enumitem}
\usepackage{tcolorbox}
\usepackage{adjustbox}
\usepackage[normalem]{ulem}

\def\eg{\textit{e.g.}}
\def\etc{\textit{etc}}

\newcommand{\myPara}[1]{\vspace{.05in}\noindent\textbf{#1.}}
\newlength\savedwidth
\newcommand\whline{\noalign{\global\savedwidth\arrayrulewidth\global\arrayrulewidth 0.8pt}\hline\noalign{\global\arrayrulewidth\savedwidth}}
\usepackage{verbatim}

\definecolor{mylb}{RGB}{229, 247, 255}
\definecolor{mygray}{gray}{.95}
\definecolor{darkred}{rgb}{0.55, 0.0, 0.0}
\definecolor{myred}{HTML}{ED058C}

\newcolumntype{a}{>{\columncolor{mylb}}c}

\title{MambaOut: Do We Really Need Mamba for Vision?}

\author{%
  Weihao Yu \quad Xinchao Wang \\
  National University of Singapore\\
  \texttt{weihaoyu@u.nus.edu \quad xinchao@nus.edu.sg} \\
  \small{Code: \url{https://github.com/yuweihao/MambaOut}}
}

\begin{document}

\maketitle
\vspace{-10mm}
\begin{center}
    \textit{In memory of Kobe Bryant}
\end{center}
\vspace{-4mm}
\begin{quote}
    ``What can I say, Mamba out.''
    \textit{--- Kobe Bryant's NBA farewell speech in 2016.}
\end{quote}

\vspace{-4mm}
\begin{figure}[H]
\hsize=\textwidth
\centering
\begin{subfigure}[b]{0.52\textwidth}
    \centering
    \includegraphics[width=1\textwidth]{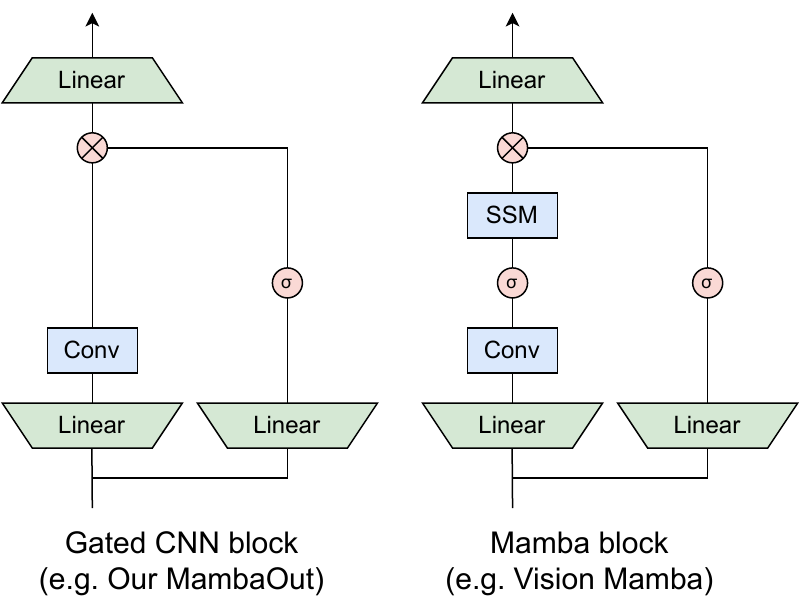}
    \vspace{-5mm}
    \caption{}
\end{subfigure}    
\begin{subfigure}[b]{0.47\textwidth}
     \centering
     \includegraphics[width=1\textwidth]{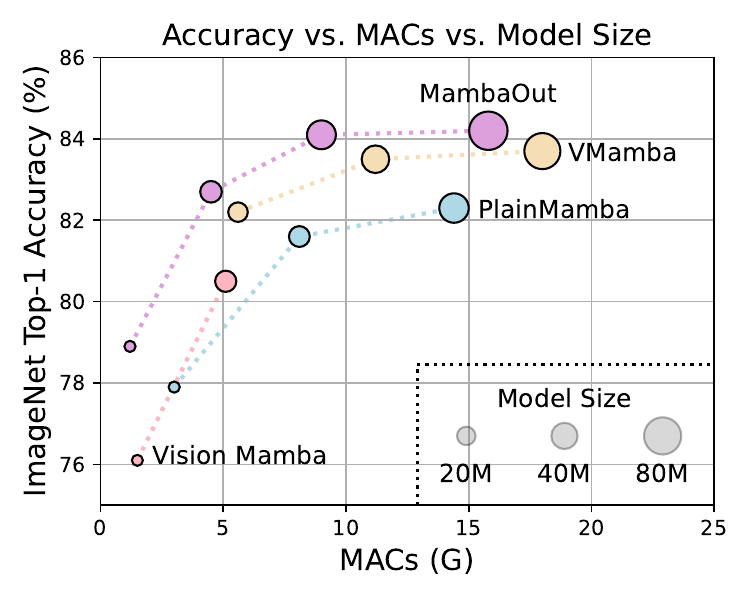}
     \vspace{-5mm}
     \caption{}
\end{subfigure}
\vspace{-5mm}
\caption{(a) Architecture of Gated CNN \cite{dauphin2017language} and Mamba \cite{gu2023mamba} blocks (omitting Normalization and shortcut). The Mamba block extends the Gated CNN with an additional state space model (SSM).
As will be conceptually discussed in Section \ref{sec:conceptual_discussion}, SSM is not necessary for image classification on ImageNet \cite{deng2009imagenet, russakovsky2015imagenet}.
To empirically verify this claim, we stack Gated CNN blocks to build a series of models named \emph{MambaOut}.
(b) MambaOut outperforms visual Mamba models, \eg, Vision Mamhba \cite{zhu2024vision}, VMamba \cite{liu2024vmamba} and PlainMamba \cite{yang2024plainmamba}, on ImageNet image classification. 
}
\label{fig:first_figure}
\end{figure}

\vspace{-5mm}
\begin{quote}
\textbf{Abstract} --- 
Mamba, an architecture with RNN-like token mixer of state space model (SSM), was recently introduced to address the quadratic complexity of the attention mechanism and subsequently applied to vision tasks\footnote{The vision tasks we discuss in this paper include image classification on ImageNet \cite{deng2009imagenet,russakovsky2015imagenet}, object detection \& instance segmentation on COCO \cite{lin2014microsoft} and semantic segmentation on ADE20K \cite{zhou2017scene}.}.
Nevertheless, the performance of Mamba for vision is often underwhelming when compared with convolutional and attention-based models. In this paper, we delve into the essence of Mamba, and conceptually conclude that Mamba is ideally suited for tasks with \emph{long-sequence} and \emph{autoregressive} characteristics. For vision tasks, as image classification does not align with either characteristic, we hypothesize that Mamba is not necessary for this task; Detection and segmentation tasks are also not \emph{autoregressive}, yet they adhere to the \emph{long-sequence} characteristic, so we believe it is still worthwhile to explore Mamba's potential for these tasks. To empirically verify our hypotheses, we construct a series of models named \emph{MambaOut} through stacking Mamba blocks while removing their core token mixer, SSM. Experimental results strongly support our hypotheses. Specifically, our MambaOut model surpasses all visual Mamba models on ImageNet image classification, indicating that Mamba is indeed unnecessary for this task. As for detection and segmentation, MambaOut cannot match the performance of state-of-the-art visual Mamba models, demonstrating the potential of Mamba for long-sequence visual tasks.

\end{quote}

\section{Introduction}
In recent years, Transformer \cite{vaswani2017attention} has become the mainstream backbone for various tasks, underpinning numerous prominent models such as BERT \cite{devlin2018bert}, GPT series \cite{radford2018improving,radford2019language,brown2020language,achiam2023gpt} and ViT \cite{dosovitskiy2020image}.
However, the token mixer of Transformer, attention \cite{bahdanau2014neural}, incurs a quadratic complexity with respect to sequence length, posing major challenges for long sequences. To address this issue, a variety of token mixers with linear complexity to token length have been introduced \cite{tay2022efficient}, such as dynamic convolution \cite{wu2019pay,wu2020lite,jiang2020convbert}, Linformer \cite{wang2020linformer}, Longformer \cite{beltagy2020longformer}, Big Bird \cite{zaheer2020big}, and Performer \cite{choromanski2020masked}.
More recently, a new wave of RNN-like models has emerged \cite{katharopoulos2020transformers,zhai2021attention,gu2021efficiently,peng2023rwkv,gu2023mamba}, drawing significant interest from the community for their capability of parallelizable training
and performing efficient inference on long sequences. Notably, models like RWKV \cite{peng2023rwkv} and Mamba \cite{gu2023mamba} are proven to be effective as the backbone for large language models (LLMs) \cite{peng2023rwkv, lieber2024jamba}.

Motivated by the promising capabilities of RNN-like models, 
various research endeavors have attempted to 
introduce Mamba \cite{gu2023mamba} into visual recognition tasks, 
exemplified by
the pioneering works of Vision Mamba \cite{zhu2024vision}, VMamba \cite{liu2024vmamba}, LocalMamba \cite{huang2024localmamba}, and PlainMamba \cite{yang2024plainmamba}, \etc. The token mixer of Mamba is the structured state space models (SSM) \cite{gu2021combining, gu2021efficiently, gu2023mamba}, under the spirit of RNN. Nevertheless, their experiments show that the SSM based models for vision, in reality, lead to underwhelming performance compared with state-of-the-art convolutional \cite{liu2022convnet,ding2022scaling,guo2023visual,rao2022hornet,yang2022focal,liu2022more,yu2024metaformer,hou2022conv2former,wang2023internimage, yu2024inceptionnext} and attention-based models \cite{dai2021coatnet,touvron2021going,dong2022cswin,yuan2022volo,li2022mvitv2,tu2022maxvit,si2022inception,yu2024metaformer}. 
This  gives rise to a compelling research question: 
\emph{Do we really need Mamba for Vision?}

In this paper, we investigate the nature of Mamba, and conceptually summarize that Mamba is ideally suited for tasks with two key characteristics: \emph{long-sequence} and \emph{autoregressive}, because of the inherent RNN mechanism of SSM~\cite{gu2021combining, gu2021efficiently, gu2023mamba} (see explanation of Figure \ref{fig:memory} and Figure \ref{fig:two_modes}). Unfortunately, not many vision tasks possess both characteristics. Image classification on ImageNet, for example, conforms to neither, while object detection \& instance segmentation on COCO and semantic segmentation on ADE20K conform only to the \emph{long-sequence}. \emph{Autoregressive} characteristic, on the other hand, demands that each token aggregate information solely from preceding and current tokens, a concept denoted as \emph{causal mode} for token mixing \cite{raffel2020exploring} (see Figure \ref{fig:two_modes}(a)). In fact, all visual recognition tasks fall within the understanding domain rather than the generative one, meaning that the model can see the entire image at once. As such, imposing additional causal constraints on token mixing in visual recognition models could lead to a performance drop (see Figure \ref{fig:two_modes}(b)). Although this issue can be mitigated via bidirectional branches \cite{schuster1997bidirectional}, it is inevitable that the issue persists within each branch.

Based on the conceptual discussion above, we propose the two hypotheses as follows: 
\begin{itemize}
    \item \textit{Hypothesis 1}: SSM is not necessary for image classification, since this task conforms to neither the \emph{long-sequence} or \emph{autoregressive} characteristic.
    \item \textit{Hypothesis 2}: SSM may be potentially beneficial for object detection \& instance segmentation and semantic segmentation, since they follow the \emph{long-sequence} characteristic, though they are not \emph{autoregressive}.
\end{itemize}

To experimentally validate our hypotheses, we developed a series of models termed \emph{MambaOut} through stacking Gated CNN \cite{dauphin2017language} blocks. The key distinction between Gated CNN and Mamba blocks lies in the existence of SSM, as illustrated in Figure \ref{fig:first_figure}(a). Experimental results demonstrate that the simpler MambaOut model, in reality, already surpasses the performance of visual Mamba models \cite{zhu2024vision, liu2024vmamba, huang2024localmamba, yang2024plainmamba}, which in turn verifies our \textit{Hypothesis~1}. We also show empirical results that
MambaOut falls short of matching the performance of state-of-the-art visual Mamba models \cite{liu2024vmamba, huang2024localmamba} in detection and segmentation tasks (see Tables \ref{tab:mask_rcnn} and \ref{tab:upernet}), which underscores the potential of SSM on these tasks and effectively validates our \textit{Hypothesis~2}.

The contributions of our paper are threefold. Firstly, we analyze the RNN-like mechanism of SSM and conceptually conclude that Mamba is suited for tasks with long-sequence and autoregressive characteristics. Secondly, we examine the characteristics of visual tasks and hypothesize that SSM is unnecessary for image classification on ImageNet since this task does not meet either characteristic, yet exploring the potential of SSM for detection and segmentation tasks remains valuable since these tasks conform to long-sequence characteristic, though they are not autoregressive. Thirdly, we develop a series of models named MambaOut based on Gated CNN blocks but without SSM. Experiments show that MambaOut effectively surpasses visual Mamba models in ImageNet image classification but does not reach the performance of state-of-the-art visual Mamba models in detection and segmentation tasks. These observations, in turn, validate our hypotheses. As such, MambaOut, because of its \emph{Occam's razor} nature, may readily serve as a natural baseline for future research on visual Mamba models.

\section{Related work}
Transformer has been widely utilized across various domains, including BERT \cite{devlin2018bert} and GPT series \cite{radford2018improving,radford2019language,brown2020language,achiam2023gpt} in NLP and ViT \cite{dosovitskiy2020image} in computer vision. However, the attention module in Transformers scales quadratically with sequence length, presenting a significant computational challenge. Numerous studies \cite{tay2022efficient} have explored various strategies to mitigate this issue, including low-rank approaches \cite{wang2020linformer}, kernelization \cite{katharopoulos2020transformers,choromanski2020masked}, token mixing range limitation \cite{beltagy2020longformer, zaheer2020big, liu2021swin, hassani2023neighborhood}, and history memory compression \cite{rae2019compressive}. More recently, RNN-like methods \cite{dai2019transformer, katharopoulos2020transformers, zhai2021attention}, particularly RWKV \cite{peng2023rwkv} and Mamba \cite{gu2023mamba}, have garnered attention for their promising results in large language models \cite{peng2023rwkv,lieber2024jamba}.

Eager exploratory researchers have quickly moved to incorporate SSM and Mamba \cite{gu2023mamba} into visual recognition tasks \cite{zhu2024vision, liu2024vmamba, huang2024localmamba, yang2024plainmamba, li2024mamba, patro2024simba, pei2024efficientvmamba, zhang2024survey, xu2024survey}.  For instance, Vision Mamba \cite{zhu2024vision} integrates Mamba \cite{gu2023mamba} to develop isotropic vision models akin to ViT \cite{dosovitskiy2020image}; VMamba \cite{liu2024vmamba} employs Mamba to construct hierarchical vision models similar to AlexNet \cite{krizhevsky2012imagenet} and ResNet \cite{he2016deep}; LocalMamba \cite{huang2024localmamba} enhances visual Mamba models \cite{zhu2024vision, liu2024vmamba} by incorporating local inductive biases; PlainMamba \cite{yang2024plainmamba} aims to further enhance the performance of isotropic Mamba models; EfficientVMamba \cite{pei2024efficientvmamba} focuses on efficiency through the introduction of atrous selective scan for lightweight visual Mamba models.

Unlike these initiatives, our work does not aim to design new visual Mamba models. Instead, we explore a pertinent research question about the necessity of Mamba \cite{gu2023mamba} in visual recognition contexts \cite{deng2009imagenet,russakovsky2015imagenet,lin2014microsoft,zhou2017scene}. We hope this paper can provide insights for future research on visual Mamba models.
\section{Conceptual discussion}
\label{sec:conceptual_discussion}
In this section, we first discuss what characteristics of tasks the Mamba model is suited for. Next, we examine whether visual recognition tasks conform to these characteristics. Based on the examination results, we propose hypotheses regarding the necessity of Mamba for vision.

\subsection{What tasks is Mamba suitable for?}
\label{sec:task_cha}
\begin{figure}[t]
  \centering
   \includegraphics[width=0.85\linewidth]{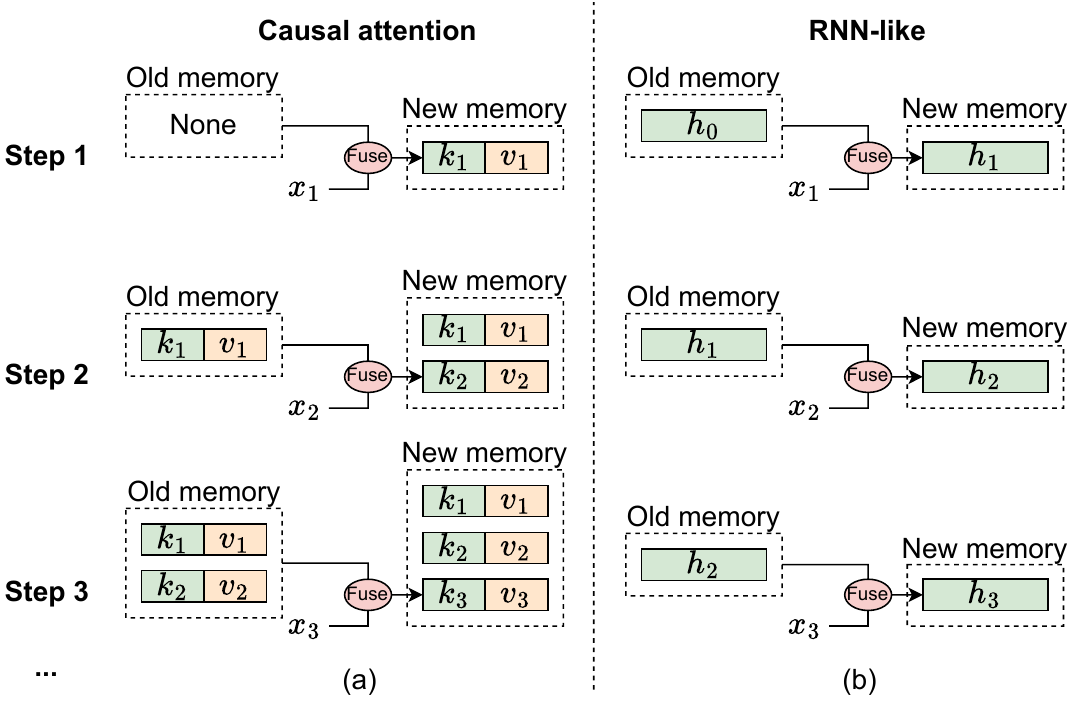}
   \caption{The mechanism illustration of causal attention and RNN-like models from memory perspective, where $x_i$ denotes the input token of $i$-th step. (a) Causal attention stores all previous tokens' keys $k$ and values $v$ as memory. The memory is updated by continuously adding the current token's key and value, so the memory is lossless, but the downside is that the computational complexity of integrating old memory and current tokens increases as the sequence lengthens.  
   Therefore, attention can effectively manage short sequences but may encounter difficulties with longer ones.
   (b) In contrast, RNN-like models compress previous tokens into fixed-size hidden state $h$, which serves as the memory. This fixed size means that RNN memory is inherently lossy, which cannot directly compete with the lossless memory capacity of attention models.  Nonetheless, \textbf{RNN-like models can demonstrate distinct advantages in processing long sequences,  as the complexity of merging old memory with current input remains constant, regardless of sequence length.}}
   \label{fig:memory}
\end{figure}

\begin{figure}[t]
\hsize=\textwidth
\centering
\begin{subfigure}[b]{0.58\textwidth}
    \centering
    \includegraphics[width=1\textwidth]{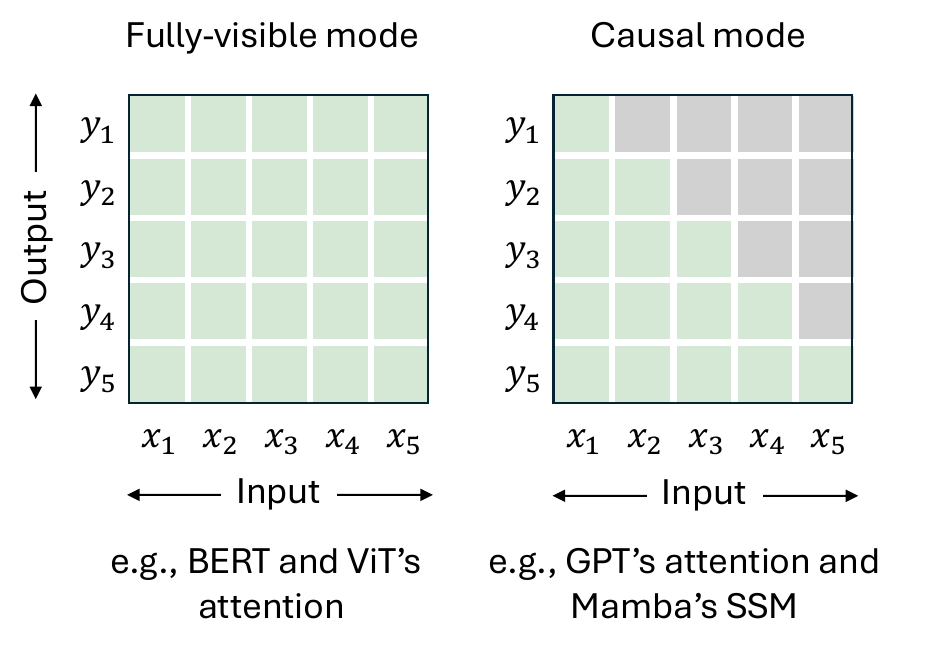}
    \caption{}
\end{subfigure}    
\begin{subfigure}[b]{0.4\textwidth}
     \centering
     \includegraphics[width=1\textwidth]{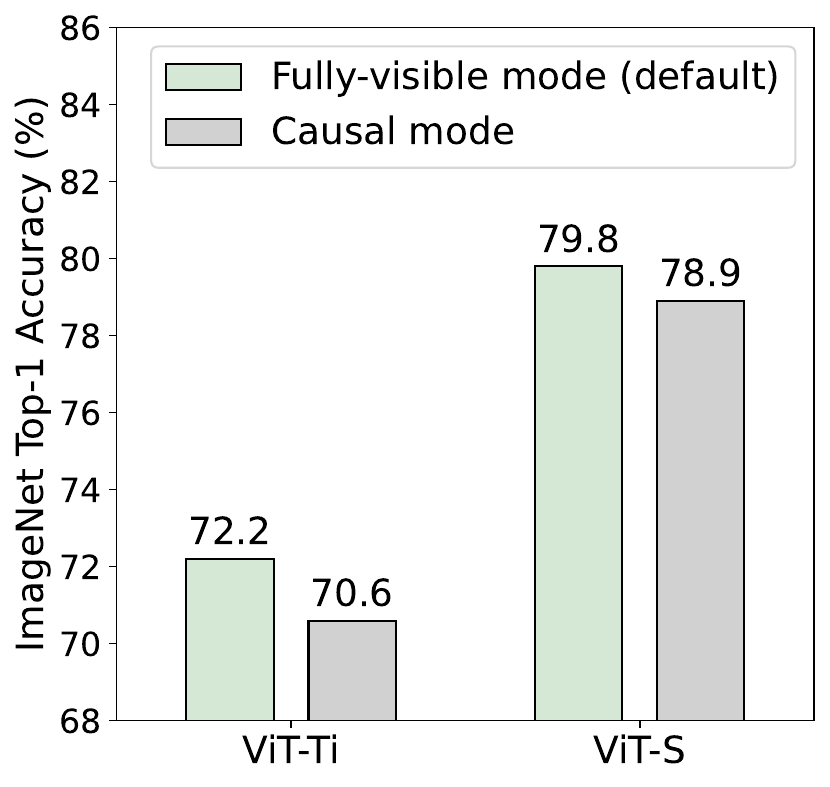}
     \caption{}
\end{subfigure}
\caption{
(a) Two modes of token mixing \cite{raffel2020exploring}. For a total of $T$ tokens, the fully-visible mode allows token $t$ to aggregate inputs from all tokens, i.e., $\{xi\}_{i=1}^{T}$, to compute its output $y_t$. In contrast, the causal mode restricts token $t$ to only aggregate inputs from preceding and current tokens $\{x_i\}_{i=1}^{t}$. By default, attention operates in fully-visible mode but can be adjusted to causal mode with causal attention masks. RNN-like models, such as Mamba's SSM \cite{gu2023mamba, gu2021efficiently}, inherently operate in causal mode due to their recurrent nature. (b) \textbf{We modify the ViT's attention \cite{dosovitskiy2020image, touvron2021training} from fully-visible to causal mode and observe performance drop on ImageNet, which indicates causal mixing is unnecessary for understanding tasks.}
}
\label{fig:two_modes}
\end{figure}

The token mixer of Mamba is selective SSM \cite{gu2021efficiently, gu2023mamba} which defines four input-dependent parameters $(\Delta, \mathbf{A}, \mathbf{B}, \mathbf{C})$ and transforms them to $(\overline{\mathbf{A}}, \overline{\mathbf{B}}, C)$ by
\begin{equation}
    \overline{\mathbf{A}} = \mathrm{exp}(\Delta A), \quad 
    \overline{\mathbf{B}} = (\Delta \mathbf{A})^{-1}(\mathrm{exp}(\Delta \mathbf{A}) - \mathbf{I}) \cdot \Delta \mathbf{B}.
\end{equation}
Then the sequence-to-sequence transformation of SSM can be expressed by
\begin{align}
    h_t & = \overline{\mathbf{A}}h_{t-1} + \overline{\mathbf{B}}x_t, \label{eq:transform}\\
    y_t & = \mathbf{C}h_t,
\end{align}
where $t$ denotes the timestep, $x_t$ represents the input, $h_t$ signifies the hidden state, and $y_t$ indicates the output. The recurrent property \cite{hochreiter1997long} of Equation \ref{eq:transform} distinguishes RNN-like SSM from causal attention. The hidden state $h$ can be seen as a fixed-size memory that stores all historical information.
Through Equation \ref{eq:transform}, this memory is updated while its size remains constant. The fixed size means the memory is inevitably lossy, but it ensures that the computational complexity of integrating the memory with the current input remains constant. 
Conversely, causal attention stores all keys and values from previous tokens as its memory, which expands by adding the current token's key and value with each new input. 
This memory is theoretically lossless. However, as more tokens are inputted, the memory size grows, thereby increasing the complexity of integrating the memory with the current input. 
The differences in memory mechanisms between RNN-like models and causal attention are further illustrated in Figure \ref{fig:memory}.

Because SSM's memory is inherently lossy, it logically falls short of the lossless memory of attention. Consequently, Mamba cannot showcase its strengths in handling short sequences, an area where attention performs well with ease. However, in scenarios involving long sequences, attention will falter due to its quadratic complexity. In this case, Mamba can distinctly highlight its efficiency in merging memory with the current input, thus managing long sequences smoothly. Therefore, Mamba is particularly well-suited for processing long sequences.

Although the recurrent nature of SSM (Equation \ref{eq:transform}) allows Mamba to handle long sequences efficiently, it introduces a significant limitation: $h_t$ can only access information from the previous and current timesteps. As illustrated in Figure \ref{fig:two_modes}, this type of token mixing is termed causal mode, which can be formulated as:
\begin{equation}
\label{eq:causal}
    y_t = f(x_1, x_2, ..., x_t),
\end{equation}
where $x_t$ and $y_t$ represent the input and output of the $t$-th token, respectively. Due to its causal nature, this mode is well-suited for autoregressive generation tasks.

Another mode is called fully-visible mode,  where each token can aggregate information from all preceding and subsequent tokens. This means the output of each token depends on the inputs from all tokens:
\begin{equation}
    y_t = f(x_1, x_2, ..., x_t, ..., x_T),
\end{equation}
where $T$ represents the total number of tokens.
The fully-visible mode is suitable for understanding tasks, where all inputs can be accessed by the model at once.

Attention is in fully-visible mode by default, but it can easily turn into causal mode by applying causal masks to the attention maps. RNN-like models inherently operate in causal mode due to their recurrent properties, as illustrated by Mamba's Equation \ref{eq:transform}. Due to this inherent characteristic, RNN-like models cannot be transformed into fully-visible mode. 
Although RNNs can approximate a fully-visible mode using bidirectional branches, each branch still individually remains in causal mode. Therefore, Mamba is well-suited for tasks that require causal token mixing, due to the inherent limitations of its recurrent properties.

In summary, Mamba is ideally suited for tasks that display the following characteristics:
\begin{itemize}
    \item \textit{Characteristic 1}: The task involves processing long sequences.
    \item \textit{Characteristic 2}: The task requires causal token mixing mode.
\end{itemize}

Next, we will discuss whether visual recognition tasks exhibit these two characteristics.

\subsection{Do visual recognition tasks have very long sequences?}
\label{sec:long_seq}

In this subsection, we explore whether visual recognition tasks necessitate long sequence modeling. We use the Transformer model \cite{vaswani2017attention} as a case study to facilitate our analysis. Consider a Transformer block with a common MLP ratio of 4; assuming its input $X \in \mathbb{R}^{L \times D}$ has a token length of $L$ and channel (embedding) dimensions of $D$, the FLOPs for the block can be calculated as:
\begin{equation}
\mathrm{FLOPs} = 24D^2L + 4DL^2.
\end{equation}

From this, we derive the ratio of the quadratic term to the linear term in $L$ as:
\begin{equation}
r_L = \frac{4DL^2}{24D^2L} = \frac{L}{6D}.
\end{equation}
If $L > 6D$, the computational load of the quadratic term in $L$ surpasses that of the linear term. This provides a simple metric to determine if the task involves long sequences. For instance, with 384 channels in ViT-S, the threshold $\tau_\mathrm{small} = 6 \times 384 = 2304$, and for 768 channels in ViT-B, $\tau_\mathrm{base} = 6 \times 768 = 4608$.

For image classification on ImageNet, the typical input image size is $224^2$, resulting in $14^2 = 196$ tokens with patch size of $16^2$. Clearly, $196$ is much less than both $\tau_\mathrm{small}$ and $\tau_\mathrm{base}$, indicating that image classification on ImageNet does not qualify as a long-sequence task.

For object detection \& instance segmentation on COCO, with an inference image size of $800 \times 1280$, and for semantic segmentation on ADE20K, with an inference image size of $512 \times 2048$, the number of tokens is approximately 4K, given patch size of $16^2$. Since $4K > \tau_\mathrm{small}$ and $4K \approx \tau_\mathrm{base}$, both detection on COCO and segmentation on ADE20K can be considered long-sequence tasks.

\subsection{Do visual recognition tasks need causal token mixing mode?}
\label{sec:causal}
As discussed in Section \ref{sec:task_cha} and illustrated in Figure \ref{fig:two_modes}, the fully-visible token mixing mode allows unrestricted range of mixing, whereas the causal mode limits the current token to only access information from preceding tokens.
Visual recognition is categorized as an understanding task, wherein the model can see the entire image at once, eliminating the need for restrictions on token mixing. Imposing additional constraints on token mixing can potentially degrade model performance. As demonstrated in Figure \ref{fig:two_modes}(b), when causal restrictions are applied to Vision Transformers (ViT) \cite{dosovitskiy2020image,touvron2021training}, a noticeable decline in performance is observed.
Generally, the fully-visible mode is appropriate for understanding tasks, while the causal mode is better suited for autoregressive tasks. This claim can also be substantiated by the observation that BERT \cite{devlin2018bert} and ViT \cite{dosovitskiy2020image} (BEiT \cite{bao2021beit} and MAE \cite{he2022masked}) are used more for understanding tasks than GPT-1/2 \cite{radford2018improving, radford2019language} and image GPT \cite{chen2020generative}. Therefore, visual recognition tasks do not need causal token mixing mode.

\subsection{Hypotheses regarding the necessity of Mamba for vision}

Based on our preceding discussion, we summarize our hypotheses regarding the necessity of introducing Mamba for visual recognition tasks as follows:
\begin{itemize}
    \item \textit{Hypothesis 1}: It is not necessary to introduce SSM for image classification on ImageNet, as this task does not meet \textit{Characteristic 1} or \textit{Characteristic 2}.
    \item \textit{Hypothesis 2}: It is still worthwhile to further explore the potential of SSM for visual detection and segmentation since these tasks align with \textit{Characteristic 1}, despite not fulfilling \textit{Characteristic 2}.
\end{itemize}

\section{Experimental verification}
\begin{figure}[t]
  \centering
   \includegraphics[width=1\linewidth]{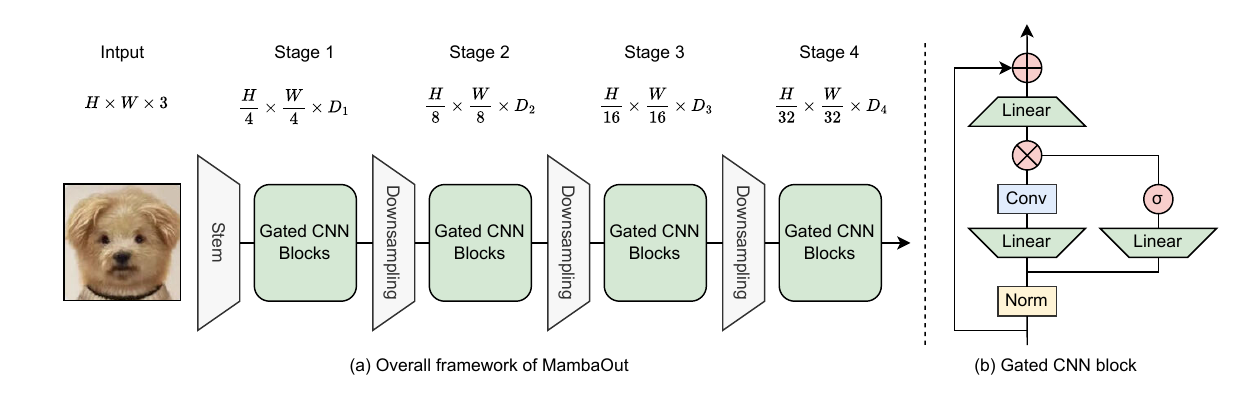}
   \caption{\textbf{(a) The overall framework of MambaOut for visual recognition.} Similar to ResNet \cite{he2016deep}, MambaOut adopts hierarchical architecture with four stages. $D_i$ represents the channel dimensions at the $i$-th stage. \textbf{(b) The architecture of Gated CNN block.} The difference between the Gated CNN block \cite{dauphin2017language} and the Mamba block \cite{gu2023mamba} lies in the absence of the SSM (state space model) in the Gated CNN block.}
   \label{fig:gated_cnn_framework}
\end{figure}

\subsection{Gated CNN and MambaOut}
Next, we aim to validate our hypotheses empirically. As depicted in Figure \ref{fig:first_figure}(a), Mamba block \cite{gu2023mamba} is based on the Gated CNN block \cite{dauphin2017language}. The meta-architecture of Gated CNN and Mamba can be considered as a simplified integration of the MetaFormer's \cite{yu2022metaformer} token mixer and an MLP, akin to MetaNeXt \cite{yu2024inceptionnext}. Formally, given the input $X \in \mathbb{R}^{N \times D}$, the meta-architecture is formulated as:
\begin{align}
X' &= \mathrm{Norm}(X), \\
Y &= (\mathrm{TokenMixer}(X'W_1) \odot \sigma (X'W_2))W_3 + X,
\end{align}
where $\mathrm{Norm}(\cdot)$ represents normalization \cite{ioffe2015batch, ba2016layer, wu2018group}; $\mathrm{TokenMixer}(\cdot)$ refers to the module to conduct token mixing \cite{yu2024metaformer}; $W_1 \in \mathbb{R}^{D \times rD}$, $W_2 \in \mathbb{R}^{D \times rD}$ and $W_3 \in \mathbb{R}^{rD \times D}$ are learnable parameters with MLP expansion $r$; $\sigma$ is activation function \cite{fukushima1969visual, hendrycks2016gaussian}. Token mixers of Gated CNN and Mamba are:
\begin{align}
    \mathrm{TokenMixer}_\mathrm{Gated CNN}(Z) &= \mathrm{Conv}(Z) \label{eq:gated_cnn} \\
    \mathrm{TokenMixer}_\mathrm{Mamba}(Z) &= \mathrm{SSM}(\sigma (\mathrm{Conv}(Z))) \label{eq:mamba}
\end{align}

Comparing Equations \ref{eq:gated_cnn} and \ref{eq:mamba}, and referencing Figure \ref{fig:first_figure}(a), the primary distinction between the Gated CNN \cite{radford2018improving} and the Mamba block \cite{gu2023mamba} lies in the presence of SSM. This prompts us to develop a series of models, termed MambaOut, which are based on the Gated CNN block without SSM. MambaOut will help us assess the necessity of Mamba for visual recognition tasks.

Specifically, we specify the token mixer of Gated CNN as depthwise convolution \cite{chollet2017xception} of $7\times 7$ kernel size, following ConvNeXt \cite{liu2022convnet,mamalet2012simplifying}. Besides, to improve the practical speed, we only conduct depthwise convolution on partial channels \cite{ma2018shufflenet, yu2024inceptionnext, chen2023run}, following InceptionNeXt \cite{yu2024inceptionnext}. As shown in Algorithm \ref{alg:gated_cnn_block},  the implementation of Gated CNN block is simple and elegant. Similar to ResNet, we adopt 4-stage framework to build MambaOut by stacking Gated CNN blocks at each stage,  as depicted in Figure \ref{fig:gated_cnn_framework}. The configuration details of each model size are shown in Table \ref{tab:mambaout_config} in the appendix.

\begin{algorithm}[t]
\caption{PyTorch code of Gated CNN block}
\label{alg:gated_cnn_block}
\definecolor{codeblue}{rgb}{0.25,0.5,0.5}
\definecolor{codekw}{rgb}{0.85, 0.18, 0.50}
\lstset{
  backgroundcolor=\color{white},
  basicstyle=\fontsize{7.5pt}{7.5pt}\ttfamily\selectfont,
  columns=fullflexible,
  breaklines=true,
  captionpos=b,
  commentstyle=\fontsize{7.5pt}{7.5pt}\color{codeblue},
  keywordstyle=\fontsize{7.5pt}{7.5pt}\color{codekw},
}
\begin{lstlisting}[language=python]
import torch
import torch.nn as nn

class GatedCNNBlock(nn.Module):
    def __init__(self, dim, expension_ratio=8/3, kernel_size=7, conv_ratio=1.0,
                 norm_layer=partial(nn.LayerNorm,eps=1e-6), 
                 act_layer=nn.GELU,
                 drop_path=0.):
        super().__init__()
        self.norm = norm_layer(dim)
        hidden = int(expension_ratio * dim)
        self.fc1 = nn.Linear(dim, hidden * 2)
        self.act = act_layer()
        conv_channels = int(conv_ratio * dim)
        self.split_indices = (hidden, hidden - conv_channels, conv_channels)
        self.conv = nn.Conv2d(conv_channels, conv_channels, kernel_size=kernel_size, padding=kernel_size//2, groups=conv_channels)
        self.fc2 = nn.Linear(hidden, dim)

    def forward(self, x):
        shortcut = x # [B, H, W, C] = x.shape
        x = self.norm(x)
        g, i, c = torch.split(self.fc1(x), self.split_indices, dim=-1)
        c = c.permute(0, 3, 1, 2) # [B, H, W, C] -> [B, C, H, W]
        c = self.conv(c)
        c = c.permute(0, 2, 3, 1) # [B, C, H, W] -> [B, H, W, C]
        x = self.fc2(self.act(g) * torch.cat((i, c), dim=-1))
        return x + shortcut
\end{lstlisting}
\end{algorithm}

\begin{table}[h!]
\renewcommand{\arraystretch}{1.3}
    \caption{
    \textbf{Performance of models on ImageNet at the resolution of $224^2$.} Our MambaOut model employs the Gated CNN block \cite{radford2018improving}. The Mamba block \cite{gu2023mamba}, derived from the Gated CNN block, incorporates an additional SSM (state space model). It is evident that visual Mamba models fall short of MambaOut's performance, let alone surpassing state-of-the-art convolutional or convolution-attention-hybrid models. *Note that VMambaV9 modifies the meta-architecture of the Mamba block to MetaFormer \cite{yu2024metaformer}, different from other visual Mamba models and MambaOut.
    }
    \label{tab:imagenet}
    \centering
    \begin{minipage}{.49\textwidth}
\scriptsize
\centering
\setlength{\tabcolsep}{4pt}
\begin{tabular}{l | c | c | c c }
\whline
\multirow{3}{*}{\makecell[c]{Model}}    & \multirow{3}{*}{\makecell[c]{Token \\ Mixing \\ Type}}     & \multirow{3}{*}{\makecell[c]{Param \\ (M)}}   & \multicolumn{2}{c}{Test@$224^2$} \\
\cline{4-5}
~ & ~ & ~ &  \multirow{2}{*}{\makecell[c]{MAC \\ (G)}} & \multirow{2}{*}{\makecell[c]{Acc \\ (\%)}} \\ 
~ & ~ & ~ & ~ & ~ \\
\whline
VAN-B0 \cite{guo2023visual} & Conv & 4 & 0.9 & 75.4 \\
MogaNet-T \cite{li2024moganet} & Conv & 5 & 1.1 & \textbf{79.0} \\
FasterNet-T1 \cite{chen2023run} & Conv & 8 & 0.9 & 76.2 \\
InceptionNeXt-A \cite{yu2024inceptionnext} & Conv & 4 & 0.5 & 75.3 \\
DeiT-Ti \cite{touvron2021training} & Attn & 6 & 1.3 & 72.2 \\
T2T-ViT-7 \cite{yuan2021tokens} & Attn & 4 & 1.1 & 71.7 \\
PVTv2-B0 \cite{wang2022pvt} & Conv + Attn & 3 & 0.6 & 70.5 \\ 
MobileViTv3-XS \cite{wadekar2022mobilevitv3} & Conv + Attn & 3 & 0.9 & 76.7 \\
EMO-6M \cite{zhang2023rethinking} & Conv + Attn & 6 & 1.0 & \textbf{79.0} \\
\hdashline
Vim-Ti \cite{zhu2024vision} & Conv + SSM & 7 & 1.5 & 76.1 \\
LocalVim-T \cite{huang2024localmamba} & Conv + SSM & 8 & 1.5 & 76.2 \\
EfficientVMamba-T \cite{pei2024efficientvmamba} & Conv + SSM & 6 & 0.8 & 76.5 \\
EfficientVMamba-S \cite{pei2024efficientvmamba} & Conv + SSM & 11 & 1.3 & 78.7 \\
MambaOut-Femto & Conv & 7 & 1.2 & \textbf{78.9} \\
\whline
PoolFormer-S24 \cite{yu2022metaformer} & Pool & 21 & 3.4 & 80.3 \\
ConvNeXt-T \cite{liu2022convnet} &  Conv   & 29 & 4.5 & 82.1  \\
VAN-B2 \cite{guo2023visual} &  Conv & 27 & 5.0 & 82.8  \\
ConvFormer-S18 \cite{yu2024metaformer} &  Conv  & 27 & 3.9 & 83.0 \\
MogaNet-S \cite{li2024moganet} & Conv & 25 & 5.0 & 83.4 \\
InternImage-T \cite{wang2023internimage} & Conv & 30 & 5 & 83.5 \\
InceptionNeXt-T \cite{yu2024inceptionnext} & Conv & 28 & 4.2 & 82.3 \\
DeiT-S \cite{touvron2021training} &  Attn  & 22 & 4.6 & 79.8  \\
T2T-ViT-14 \cite{yuan2021tokens} &  Attn   & 22 & 4.8 & 81.5 \\
Swin-T \cite{liu2021swin} &  Attn   & 29 & 4.5 & 81.3  \\
Focal-Tiny \cite{yang2021focal} & Attn & 29 &  4.9 &  82.2 \\
CSWin-T \cite{dong2022cswin} &  Attn  & 23 & 4.3 & 82.7  \\
CoAtNet-0 \cite{dai2021coatnet} &  Conv + Attn  & 25 & 4.2 & 81.6 
 \\
iFormer-S \cite{si2022inception} &  Conv + Attn & 20 & 4.8 & 83.4 \\
MOAT-0 \cite{yang2023moat} & Conv + Attn & 28 &  5.7 & 83.3 \\
CAFormer-S18  \cite{yu2024metaformer} &  Conv + Attn  & 26 & 4.1 & 83.6 \\
SG-Former-S \cite{ren2023sg} & Conv + Attn & 23 & 4.8 & 83.2 \\
TransNeXt-Tiny \cite{shi2023transnext} & Conv + Attn & 28 & 5.7 &  \textbf{84.0} \\
\hdashline
Vim-S \cite{zhu2024vision} & Conv + SSM & 26 & 5.1 & 80.5 \\
VMamba-T \cite{liu2024vmamba} & Conv + SSM & 22 & 5.6 & 82.2 \\
Mamba-2D-S \cite{li2024mamba} & Conv + SSM & 24 & -- & 81.7 \\
LocalVim-S \cite{huang2024localmamba} & Conv + SSM & 28 & 4.8 & 81.2 \\
LocalVMamba-T \cite{huang2024localmamba} & Conv + SSM & 26 & 5.7 & 82.7 \\
EfficientVMamba-B \cite{pei2024efficientvmamba} & Conv + SSM & 33 & 4.0 & 81.8 \\
PlainMamba-L1 \cite{yang2024plainmamba} & Conv + SSM & 7 & 3.0 & 77.9 \\
VMambaV9-T* \cite{liu2024vmamba} & Conv + SSM & 31 & 4.9 & 82.5 \\
MambaOut-Tiny & Conv & 27 & 4.5 & \textbf{82.7} \\
\whline
\end{tabular}
\end{minipage}
\begin{minipage}{.49\textwidth}
\scriptsize
\centering
\setlength{\tabcolsep}{4pt}
\begin{tabular}{l | c | c | c c }
\whline
\multirow{3}{*}{\makecell[c]{Model}}    & \multirow{3}{*}{\makecell[c]{Token \\ Mixing \\ Type}}     & \multirow{3}{*}{\makecell[c]{Param \\ (M)}}   & \multicolumn{2}{c}{Test@$224^2$} \\
\cline{4-5}
~ & ~ & ~ &  \multirow{2}{*}{\makecell[c]{MAC \\ (G)}} & \multirow{2}{*}{\makecell[c]{Acc \\ (\%)}} \\ 
~ & ~ & ~ & ~ & ~ \\
\whline
ConvNeXt-S \cite{liu2022convnet} & Conv & 50 & 8.7 & 83.1 \\
VAN-B3 \cite{guo2023visual} & Conv & 45 & 9.0 & 83.9 \\
ConvFormer-S36 \cite{yu2024metaformer} & Conv & 40 & 7.6 & 84.1 \\
InternImage-S \cite{wang2023internimage} & Conv & 50 & 8 & 84.2 \\
MogaNet-B \cite{li2024moganet} & Conv & 44 & 9.9 & 84.3 \\
T2T-ViT-19 \cite{yuan2021tokens} & Attn & 39 & 8.5 & 81.9 \\
Swin-S \cite{liu2021swin} & Attn & 50 & 8.7 & 83.0 \\
Focal-Small \cite{yang2021focal} & Attn & 51 & 9.1 & 83.5 \\
CSWin-S \cite{dong2022cswin} & Attn & 35 & 6.9 & 83.6 \\
MViTv2-S \cite{li2022mvitv2} & Attn & 35 & 7.0 & 83.6 \\
CoAtNet-1 \cite{dai2021coatnet} & Conv + Attn & 42 & 8.4 & 83.3 \\
UniFormer-B \cite{li2023uniformer} & Conv + Attn & 50 & 8.3 & 83.9 \\
CAFormer-S36 \cite{yu2024metaformer} & Conv + Attn & 39 & 8.0 & 84.5 \\
SG-Former-M \cite{ren2023sg} & Conv + Attn & 39 & 7.5 & 84.1 \\
TransNeXt-Small \cite{shi2023transnext} & Conv + Attn & 50 & 10.3 & \textbf{84.7} \\
\hdashline
VMamba-S \cite{liu2024vmamba} & Conv + SSM & 44 & 11.2 & 83.5 \\
LocalVMamba-S \cite{huang2024localmamba} & Conv + SSM & 50 & 11.4 & 83.7 \\
PlainMamba-L2 \cite{yang2024plainmamba} & Conv + SSM & 25 & 8.1 & 81.6 \\
VMambaV9-S \cite{liu2024vmamba} & Conv + SSM & 50 & 8.7 & 83.6 \\
MambaOut-Small & Conv & 48 & 9.0 & \textbf{84.1} \\
\whline
ConvNeXt-B \cite{liu2022convnet} & Conv & 89 & 15.4 & 83.8 \\
RepLKNet-31B \cite{ding2022scaling} & Conv & 79 & 15.3 & 83.5 \\
ConvFormer-M36 \cite{yu2024metaformer} & Conv & 57 & 12.8 & 84.5 \\
HorNet-B \cite{rao2022hornet} & Conv & 88 & 15.5 & 84.3 \\
MogaNet-L \cite{li2024moganet} & Conv & 83 & 15.9 & 84.7 \\
InternImage-B \cite{wang2023internimage} & Conv & 97 & 16 & 84.9 \\
DeiT-B \cite{touvron2021training} & Attn & 86 & 17.5 & 81.8 \\
T2T-ViT-24 \cite{yuan2021tokens} & Attn & 64 & 13.8 & 82.3 \\
Swin-B \cite{liu2021swin} & Attn & 88 & 15.4 & 83.5 \\
CSwin-B \cite{dong2022cswin} & Attn & 78 & 15.0 & 84.2 \\
MViTv2-B \cite{li2022mvitv2} & Attn & 52 & 10.2 & 84.4 \\
CoAtNet-2 \cite{dai2021coatnet} & Conv + Attn & 75 & 15.7 & 84.1 \\
iFormer-L \cite{si2022inception} & Conv + Attn & 87 & 14.0 & 84.8 \\
MOAT-2 \cite{yang2023moat} & Conv + Attn & 73 & 17.2 &  84.7 \\
CAFormer-M36 \cite{yu2024metaformer} & Conv + Attn & 56 & 13.2 & \textbf{85.2} \\
TransNeXt-Base \cite{shi2023transnext} & Conv + Attn & 90 & 18.4 & 84.8 \\
\hdashline
VMamba-B \cite{liu2024vmamba} & Conv + SSM & 75 & 18.0 & 83.7 \\
Mamba-2D-B \cite{li2024mamba} & Conv + SSM & 92 & -- & 83.0 \\
PlainMamba-L3 \cite{yang2024plainmamba} & Conv + SSM & 50 & 14.4 & 82.3 \\
VMambaV9-B \cite{liu2024vmamba} & Conv + SSM & 89 & 15.4 & 83.9 \\
MambaOut-Base & Conv & 85 & 15.8 & \textbf{84.2} \\
\whline
\end{tabular}
\end{minipage}

    \vspace{-3mm}
\end{table}

\subsection{Image classification on ImageNet}
\myPara{Setup} ImageNet \cite{deng2009imagenet, russakovsky2015imagenet} serves as the gold standard benchmark for image classification, encompassing a wide array of 1,000 common classes. It comprises approximately 1.3 million training images and 50,000 validation images. The training scheme follows DeiT \cite{touvron2021training} without distillation. Specifically, the used data augmentation contains random resized crop (input image size of $224^2$), horizontal flip, RandAugment \cite{cubuk2020randaugment}, Mixup \cite{zhang2018mixup}, CutMix \cite{yun2019cutmix}, Random Erasing \cite{zhong2020random} and color jitter; Regularization techniques include weight decay, stochastic depth \cite{huang2016deep} and label smoothing \cite{szegedy2016rethinking}. All our models are trained by AdamW \cite{loshchilov2017decoupled, kingma2014adam}. The learning rate scaling rule is $\mathrm{lr} = \frac{\mathrm{batch size}}{1024} \times 10^{-3}$. In this paper, we set the batch size to 4096, so the learning rate is $0.004$. Our MambaOut models are implemented with PyTorch \cite{paszke2019pytorch} and timm \cite{rw2019timm} libraries and trained on TPU v3. More training hyper-parameters are shown in Table \ref{tab:mambaout_hyperparameter} in the appendix. 

\myPara{Results} The performance of our MambaOut models, visual Mamba models, and other various convolution and attention-based models on ImageNet \cite{deng2009imagenet, russakovsky2015imagenet} is presented in Table \ref{tab:imagenet}. Notably, our MambaOut models, which do not incorporate SSM, consistently outperform visual Mamba models \cite{zhu2024vision, liu2024vmamba, huang2024localmamba, pei2024efficientvmamba, yang2024plainmamba} that include SSM across all model sizes. For instance, the MambaOut-Small model achieves top-1 accuracy of 84.1\%, 0.4\% higher than that of LocalVMamba-S \cite{huang2024localmamba}, while requiring only 79\% of the MACs. These results strongly support our \textit{Hypothesis 1}, which posits that introducing SSM for image classification on ImageNet is unnecessary, aligning with the principle of Occam's razor.

Additionally, visual Mamba models currently exhibit a significant performance gap when compared to state-of-the-art convolution and attention models. For instance, the CAFormer-M36 \cite{yu2024metaformer}, which employs old-fashioned token mixers of simple separable convolutions from MobileNetV2 \cite{sandler2018mobilenetv2} and vanilla attention from Transformer \cite{vaswani2017attention} invented more than 7 years ago, outperforms all visual Mamba models of comparable size by more than 1\% accuracy.  Should future research aim to challenge our \textit{Hypothesis 1}, it will be necessary to develop visual Mamba models with token mixers of convolution and SSM to achieve state-of-the-art performance on ImageNet.

\subsection{Object detection \& instance segmentation on COCO}
\myPara{Setup} COCO 2017 \cite{lin2014microsoft} serves as a widely recognized benchmark for object detection and instance segmentation. In our experiments, MambaOut is employed as the backbone within Mask R-CNN \cite{he2017mask}, initialized with weights pre-trained on ImageNet. We adhere to the standard 1$\times$ training schedule of 12 epochs. The training images are resized such that the shorter side measures 800 pixels, while the longer side does not exceed 1333 pixels. The AdamW optimizer \cite{loshchilov2017decoupled, kingma2014adam} is used with a learning rate of 0.0001 and a total batch size of 16. Our implementation leverages the PyTorch \cite{paszke2019pytorch} and mmdetection \cite{mmdetection} libraries. We utilize FP16 precision to save training costs. The experiments are conducted on 4 GPUs of NVIDIA 4090. 

\myPara{Results} Although MambaOut can surpass some visual Mamba models \cite{pei2024efficientvmamba,yang2024plainmamba} in object detection and instance segmentation on COCO \cite{lin2014microsoft}, it still lags behind the state-of-the-art visual Mambas, such as VMamba \cite{liu2024vmamba} and LocalVMamba \cite{liu2024vmamba}. For instance, the performance of MambaOut-Tiny as the backbone for Mask R-CNN trails VMamba-T \cite{liu2024vmamba} by 1.4 $\mathrm{AP}^{\mathrm{b}}$ and 1.1 $\mathrm{AP}^{\mathrm{m}}$. This performance disparity underscores the benefits of integrating Mamba in long-sequence visual tasks, reinforcing our \textit{Hypothesis 2}. However, visual Mamba still exhibits a significant performance gap when compared to the state-of-the-art convolution-attention-hybrid models, TransNeXt \cite{shi2023transnext}. Visual Mamba needs to further validate its effectiveness by outperforming other state-of-the-art models in the visual detection task.

\begin{table}[t]
\caption[caption]{
\textbf{Performance of object detection and instance segmentation on COCO with Mask R-CNN. } The MACs are measured with input size of $800 \times 1280$. 
}
\label{tab:mask_rcnn}
\centering
\scriptsize
\setlength{\tabcolsep}{7pt}
\begin{tabular}{lc|cc|cccccc}
\whline
\multirow{2}{*}{Backbone} & \multirow{2}{*}{\makecell[c]{Token \\ Mixing Type}} & \multirow{2}{*}{\makecell[c]{Param \\ (M)}} & \multirow{2}{*}{\makecell[c]{MAC \\(G)}} & \multicolumn{6}{c}{Mask R-CNN 1$\times$ schedule} \\
\cline{5-10}
&  &  &  &  $\text{AP}^{\text{b}}$ & $\text{AP}^{\text{b}}_{50}$ & $\text{AP}^{\text{b}}_{75}$ & $\text{AP}^{\text{m}}$ & $\text{AP}^{\text{m}}_{\text{50}}$ & $\text{AP}^{\text{m}}_{75}$ \\
\whline
ConvNeXt-T \cite{liu2022more} & Conv   & 48 & 262 & 44.2 & 66.6 & 48.3 & 40.1 & 63.3 & 42.8 \\
FocalNet-T \cite{yang2022focal} & Conv & 49 & 268 & 46.1 & 68.2 & 50.6 & 41.5 & 65.1 & 44.5 \\
Swin-T \cite{liu2021swin} & Attn & 48 & 267 & 42.7 & 65.2 & 46.8 & 39.3 & 62.2 & 42.2  \\
ViT-Adapter-S \cite{chen2022vision} & Attn & 48 & 403 & 44.7 & 65.8 & 48.3 & 39.9 & 62.5 & 42.8 \\
CSWin-T \cite{dong2022cswin} & Attn & 42 & 279 & 46.7 & 68.6 & 51.3 & 42.2 & 65.6 & 45.4 \\
PVTv2-B2 \cite{wang2022pvt} & Conv + Attn & 45 & 309 & 45.3 & 67.1 & 49.6 & 41.2 & 64.2 & 44.4 \\
SG-Former-S \cite{ren2023sg} & Conv + Attn & 41 & -- & 47.4 & 69.0 & 52.0 & 42.6 & 65.9 & 46.0  \\
TransNeXt-Tiny \cite{shi2023transnext} & Conv + Attn & 48 & 356 & \textbf{49.9} & 71.5 & 54.9 & \textbf{44.6} & 68.6 & 48.1 \\
\hdashline
VMamba-T \cite{liu2024vmamba} & Conv + SSM & 42 & 286 & 46.5 &  68.5 & 50.7 & 42.1 & 65.5 & 45.3  \\
LocalVMamba-T \cite{huang2024localmamba} & Conv + SSM & 45 & 291 & 46.7 & 68.7 & 50.8 & 42.2 & 65.7 & 45.5 \\
EfficientVMamba-B \cite{pei2024efficientvmamba} & Conv + SSM & 53 & 252 & 43.7 & 66.2 & 47.9 & 40.2 & 63.3 & 42.9 \\
VMambaV9-T \cite{liu2024vmamba} & Conv + SSM & 50 & 270 & \textbf{47.4} & 69.5 & 52.0 & \textbf{42.7} & 66.3 & 46.0 \\
PlainMamba-L1 \cite{yang2024plainmamba} & Conv + SSM & 31 & 388 & 44.1 & 64.8 & 47.9 & 39.1 & 61.6 & 41.9  \\
MambaOut-Tiny & Conv & 43 & 262 & 45.1 & 67.3 & 49.6 & 41.0 & 64.1 & 44.1 \\
\whline
ConvNeXt-S \cite{liu2022more} & Conv & 70 & 348 & 45.4 & 67.9 & 50.0 & 41.8 & 65.2 & 45.1  \\
FocalNet-S \cite{yang2022focal} & Conv & 72 & 365 &  48.3 &  70.5 & 53.1 & 43.1 & 67.4 & 46.2 \\
Swin-S \cite{liu2021swin} & Attn & 69 & 354 & 44.8 & 66.6 & 48.9 & 40.9 & 63.2 & 44.2 \\
CSWin-S \cite{dong2022cswin} & Attn & 54 & 342 & 47.9 & 70.1 & 52.6 & 43.2 & 67.1 & 46.2  \\
PVTv2-B3 \cite{wang2022pvt} & Conv + Attn & 65 & 397 & 47.0 & 68.1 & 51.7 & 42.5 & 65.7 & 45.7  \\
SG-Former-M \cite{ren2023sg} & Conv + Attn & 51 & -- & 48.2 & 70.3 & 53.1 & 43.6 & 66.9 & 47.0  \\
TransNeXt-Small \cite{shi2023transnext} & Conv + Attn & 69 & 516 & \textbf{51.1} & 72.6 & 56.2 & \textbf{45.5} & 69.8 & 49.1 \\
\hdashline
VMamba-S \cite{liu2024vmamba} & Conv + SSM & 64 & 400 & 48.2 & 69.7 & 52.5 & 43.0 & 66.6 & 46.4  \\ 
LocalVMamba-S \cite{huang2024localmamba} & Conv + SSM & 69 & 414 & 48.4 & 69.9 & 52.7 & 43.2 & 66.7 & 46.5 \\
VMambaV9-S \cite{liu2024vmamba} & Conv + SSM & 64 & 357 & \textbf{48.7} & 70.0 & 53.4 & \textbf{43.7} & 67.3 & 47.0  \\
MambaOut-Small & Conv & 65 & 354 & 47.4 & 69.1 & 52.4 & 42.7 & 66.1 & 46.2 \\
\whline
ConvNeXt-B \cite{liu2022more} & Conv & 108 & 486 & 47.0 & 69.4 & 51.7 & 42.7 & 66.3 & 46.0  \\
FocalNet-B \cite{yang2022focal} & Conv & 111 & 507 &  49.0 & 70.9 & 53.9 & 43.5 & 67.9 & 46.7 \\
Swin-B \cite{liu2021swin} & Attn & 107 & 496 & 46.9 & -- & -- & 42.3 & -- & --   \\
ViT-Adapter-B \cite{chen2022vision} & Attn & 102 & 557 & 47.0 & 68.2 & 51.4 & 41.8 & 65.1 & 44.9  \\
CSWin-B \cite{dong2022cswin} & Attn & 97 & 526 & 48.7 & 70.4 & 53.9 & 43.9 & 67.8 & 47.3 \\
PVTv2-B5 \cite{wang2022pvt} & Conv + Attn & 102 & 557 & 47.4 & 68.6 & 51.9 & 42.5 & 65.7 & 46.0 \\
TransNeXt-Base \cite{shi2023transnext} & Conv + Attn & 109 & 728 & \textbf{51.7} & 73.2 & 56.9 & \textbf{45.9} & 70.5 & 49.7  \\
\hdashline
VMamba-B \cite{liu2024vmamba} & Conv + SSM & 96 & 540 & 48.5 & 69.6 & 53.0 & 43.1 & 67.0 & 46.4 \\ 
PlainMamba-L2 \cite{yang2024plainmamba} & Conv + SSM & 53 & 542 & 46.0 & 66.9 & 50.1 & 40.6 & 63.8 & 43.6 \\
VMambaV9-B \cite{liu2024vmamba} & Conv + SSM & 108 & 485 & \textbf{49.2} & 70.9 & 53.9 & \textbf{43.9} & 67.7 & 47.6  \\
MambaOut-Base & Conv & 100 & 495 & 47.4 & 69.3 & 52.2 & 43.0 & 66.4 & 46.3  \\
\whline
\end{tabular}

\normalsize
\end{table}

\begin{table}[t]
\caption{\textbf{Performance of Semantic segmentation with UperNet \cite{xiao2018unified} on ADE20K~\cite{zhou2017scene} validation set.} The MACs are measured with input size of $512 \times 2048$.}
\label{tab:upernet}
\centering
\scriptsize
\begin{tabular}{lc|cccc}
\whline
\multirow{2}{*}{Backbone} & \multirow{2}{*}{\makecell[c]{Token \\ Mixing Type}} & \multicolumn{4}{c}{UperNet}\\
\cline{3-6}
&  & Param (M) & MAC (G) & mIoU (SS) & mIoU (MS) \\
    \whline
    ConvNeXt-T ~\cite{liu2022more} & Conv   & 60 & 939 & 46.0 & 46.7 \\
    HorNet-T \cite{rao2022hornet} & Conv & 55 & 924 &  49.2 & 49.3  \\ 
    ConvFormer-S18 \cite{yu2024metaformer} & Conv & 54 & 925 & 47.5 & 48.6 \\
    InternImage-T \cite{wang2023internimage} & Conv & 59 & 944 & 47.9 & 48.1 \\
    Swin-T~\cite{liu2021swin}  & Attn    & 60 & 945 & 44.4 & 45.8 \\
    Twins-S  \cite{chu2021twins} & Attn & 54 & 901 & 46.2 & 47.1 \\
    Focal-T \cite{yang2021focal} & Attn & 62 & 998 & 45.8 & 47.0 \\
    CSWin-T \cite{dong2022cswin} & Attn & 60 & 959 & 49.3 & 50.7 \\
    UniFormer-S \cite{li2023uniformer} & Conv + Attn & 52 & 955 &  47.0 &  48.5 \\
    CAFormer-S18 \cite{yu2024metaformer} & Conv + Attn & 54 & 1024 & 48.1 & 48.9 \\
    SG-Former-S \cite{ren2023sg} & Conv + Attn & 53 & 989 & 49.9 & 51.5  \\
    TransNeXt-Tiny \cite{shi2023transnext} & Conv + Attn & 59 & 978 & \textbf{51.1} & \textbf{51.2} \\
    \hdashline
    VMamba-T \cite{liu2024vmamba} & Conv + SSM & 55 & 964 & 47.3 & 48.3 \\
    LocalVMamba-T \cite{huang2024localmamba} & Conv + SSM & 57 & 970 & 47.9 & \textbf{49.1} \\
    EfficientVMamba-B \cite{pei2024efficientvmamba} & Conv + SSM & 65 & 930 & 46.5 & 47.3 \\
    PlainMamba-L2 \cite{yang2024plainmamba} & Conv + SSM & 55 & 285 & -- & 46.8 \\
    PlainMamba-L3 \cite{yang2024plainmamba} & Conv + SSM & 81 & 419 & -- & \textbf{49.1} \\
    VMambaV9-T \cite{liu2024vmamba} & Conv + SSM & 62 & 948 & \textbf{48.3} & 48.6 \\
    MambaOut-Tiny & Conv & 54 & 938 & 47.4 & 48.6 \\
    \whline
    ConvNeXt-S~\cite{liu2022more}  & Conv      & 82 & 1027 & 48.7 & 49.6 \\
    HorNet-S \cite{rao2022hornet} & Conv & 85 & 1027 & 50.0 & 50.5 \\ 
    ConvFormer-S36 \cite{yu2024metaformer} & Conv & 67 & 1003 & 49.6 & 50.7 \\
    InternImage-S \cite{wang2023internimage} & Conv & 80 & 1017 & 50.1 & 50.9 \\
    Swin-S~\cite{liu2021swin}  & Attn & 81 & 1038 & 47.6 & 49.5 \\
    Twins-B \cite{chu2021twins} & Attn & 89 & 1020 &  47.7  & 48.9 \\
    Focal-S \cite{yang2021focal} & Attn & 85 & 1130 & 48.0 & 50.0 \\ 
    CSWin-S \cite{dong2022cswin} & Attn & 65 & 1027 & 50.4 & 51.5 \\
    CAFormer-S36 \cite{yu2024metaformer} & Conv + Attn & 67 & 1197 & 50.6 & 50.8 \\
    SG-Former-M \cite{ren2023sg} & Conv + Attn & 68 & 1114 & 51.2 & 52.1 \\
    TransNeXt-Small \cite{shi2023transnext} & Conv + Attn & 80 & 1089 & \textbf{52.2} & \textbf{52.3} \\
    \hdashline
    VMamba-S \cite{liu2024vmamba} & Conv + SSM & 76 & 1081 & 49.5 & 50.5 \\
    LocalVMamba-S \cite{huang2024localmamba} & Conv + SSM & 81 & 1095 & 50.0 & 51.0 \\
    VMambaV9-S \cite{liu2024vmamba} & Conv + SSM & 82 & 1039 & \textbf{50.6} & \textbf{51.2}  \\
    MambaOut-Small & Conv & 76 & 1032 & 49.5 & 50.6 \\
	\whline
     ConvNeXt-B ~\cite{liu2022more} & Conv  & 122 & 1170 & 49.1 & 49.9 \\
     HorNet-B \cite{rao2022hornet} & Conv &  126 & 1171 &  50.5 & 50.9 \\
     ConvFormer-M36 \cite{yu2024metaformer} & Conv & 85 & 1113 & 50.4 & 51.3 \\
     InternImage-B \cite{wang2023internimage} & Conv & 128 & 1185 & 50.8 & 51.3 \\
    Swin-B~\cite{liu2021swin}  & Attn    & 121 & 1188 & 48.1 & 49.7 \\
    Twins-L \cite{chu2021twins} & Attn & 133 & 1164 & 48.8 & 50.2 \\
    Focal-B \cite{yang2021focal} & Attn & 126 & 1354 & 49.0 & 50.5 \\
    CSWin-B \cite{dong2022cswin} & Attn & 110 & 1222 & 51.1 & 52.2 \\
    UniFormer-B \cite{li2023uniformer} & Conv + Attn & 80 & 1106 & 49.5 & 50.7 \\
     CAFormer-M36 \cite{yu2024metaformer} & Conv + Attn & 84 & 1346 & 51.7 & 51.7 \\
     SG-Former-B \cite{ren2023sg} & Conv + Attn & 109 & 1304 & 52.0 &  52.7 \\
     TransNeXt-Base \cite{shi2023transnext} & Conv + Attn & 121 & 1268 & \textbf{53.0} & \textbf{53.4} \\
     \hdashline
     VMamba-B \cite{liu2024vmamba} & Conv + SSM & 110 & 1226 & 50.0 & 51.3 \\
     VMambaV9-B \cite{liu2024vmamba} & Conv + SSM & 122 & 1170 & \textbf{51.0} & \textbf{51.6} \\
     MambaOut-Base & Conv & 112 & 1178 & 49.6 & 51.0 \\
\whline
\end{tabular}
\normalsize
\vspace{-3mm}
\end{table}

\subsection{Semantic segmentation on ADE20K}
\myPara{Setup}
ADE20K \cite{zhou2017scene}, a widely-used benchmark for the semantic segmentation task, encompasses 150 semantic categories. It includes 20,000 images in the training set and 2,000 images in the validation set. In our experiments, Mamba is employed as the backbone for UperNet \cite{xiao2018unified}, with initialization from ImageNet pre-trained weights. The training is conducted using the AdamW optimizer \cite{kingma2014adam, loshchilov2017decoupled} with learning rate of 0.0001 and batch size of 16 for 160,000 iterations.
Our implementation utilizes the PyTorch \cite{paszke2019pytorch} and mmsegmentation \cite{mmseg2020} libraries. Experiments are performed on four GPUs of NVIDIA 4090, with FP16 precision to enhance the training speed.

\myPara{Results} The performance trend for semantic segmentation on ADE20K is similar to object detection on COCO. MambaOut can outperform some visual Mamba models but cannot match the results of state-of-the-art Mamba models. For instance, LocalVMamba-T \cite{huang2024localmamba} surpasses MambaOut-Tiny by 0.5 mIoU in both single scale (SS) and multi-scale (MS) evaluations, further corroborating our \textit{Hypothesis 2} empirically. Additionally, visual Mamba models continue to exhibit notable performance deficits when compared to the more advanced hybrid models that integrate convolution and attention mechanisms, such as SG-Former \cite{ren2023sg} and TransNeXt \cite{shi2023transnext}.  Visual Mamba needs to further showcase its long-sequence modeling strengths by delivering stronger performance in visual segmentation task.

\section{Conclusion}
In this paper, we discuss the Mamba mechanism conceptually and conclude that it is ideally suited for tasks with long-sequence and autoregressive characteristics. We analyze common visual tasks against these criteria and argue that introducing Mamba for ImageNet image classification is unnecessary, as it meets neither characteristic. However, the potential of Mamba for visual detection and segmentation tasks, which align with at least the long-sequence characteristic, merits further exploration. To substantiate our claims empirically, we develop MambaOut models that employ Mamba blocks without their core token mixer, SSM. MambaOut surpasses all visual Mamba models on ImageNet, yet it exhibits a notable performance gap compared to state-of-the-art visual Mamba models, thereby validating our assertions.  Due to computational resource limitations, this paper only verifies the Mamba concept for visual tasks. In the future, we may further explore Mamba and RNN concepts as well as the integration of RNN and Transformer for large language models (LLMs) and large multimodal models (LMMs).

\section*{Acknowledgement}
Weihao was partly supported by Snap Research Fellowship, Google TPU Research Cloud (TRC), and Google Cloud Research Credits program. We thank Dongze Lian, Qiuhong Shen, Xingyi Yang, and Gongfan Fang for valuable discussions.

{
\small
\bibliographystyle{plain}
\bibliography{references}
}

\appendix
\section{More details of MambaOut models}
The MambaOut model configurations are shown in Table \ref{tab:mambaout_config} and the hyper-parameters to train MambaOut on ImageNet are shown in Table \ref{tab:mambaout_hyperparameter}.

\begin{table}[h]
\footnotesize
\centering
\caption{ Configurations of MambaOut models. The contents in the tuples represent the configurations in the four stages of the models.
}
\begin{tabular}{l | c | c | c | c}
\whline
Size & Femto & Tiny & Small & Base \\
\whline
Stem & \multicolumn{4}{c}{$3\times 3$ conv with stride 2; Norm; GELU; $3\times 3$ conv with stride 2, Norm} \\
Downsampling layers & \multicolumn{4}{c}{$3 \times 3$ conv with stride 2} \\
Token mixer & \multicolumn{4}{c}{$7 \times 7$ depthwise conv} \\
MLP ratio & \multicolumn{4}{c}{8/3} \\
Classifier head   & \multicolumn{4}{c}{Global average pooling, Norm, MLP} \\
\hline
\# Blocks & (3, 3, 9, 3) & (3, 3, 9, 3) & (3, 4, 27, 3) & (3, 4, 27, 3) \\
\# Channel & (48, 96, 192, 288) & (96, 192, 384, 576) & (96, 192, 384, 576) & (128, 256, 512, 768) \\
\hline
Parameters (M) & 7.3 & 26.5 & 48.5 & 84.8 \\
MACs (G) & 1.2 & 4.5 & 9.0 & 15.8 \\
\whline
\end{tabular}

\label{tab:mambaout_config}
\end{table}

\begin{table}[h]
\footnotesize
\centering
\caption{Hyper-parameters of MambaOut on ImageNet image classification.}
\label{tab:mambaout_hyperparameter}
\begin{tabular}{@{}l|cccc}
\whline
 & \multicolumn{4}{c}{MambaOut} \\ 
\cline{2-5}
 & Femto & Tiny & Small & Base \\
\whline
Input resolution & \multicolumn{4}{c}{$224^2$} \\
Epochs & \multicolumn{4}{c}{300} \\
Batch size & \multicolumn{4}{c}{4096} \\
Optimizer & \multicolumn{4}{c}{AdamW} \\
Adam $\epsilon$ & \multicolumn{4}{c}{1e-8} \\
Adam $(\beta_1, \beta_2)$ & \multicolumn{4}{c}{(0.9, 0.999)} \\
Learning rate & \multicolumn{4}{c}{4e-3} \\
Learning rate decay & \multicolumn{4}{c}{Cosine} \\
Gradient clipping & \multicolumn{4}{c}{None} \\
Warmup epochs & \multicolumn{4}{c}{20} \\
Weight decay & \multicolumn{4}{c}{0.05} \\
Rand Augment & \multicolumn{4}{c}{9/0.5} \\
Repeated Augmentation & \multicolumn{4}{c}{off} \\
Cutmix & \multicolumn{4}{c}{1.0} \\
Mixup & \multicolumn{4}{c}{0.8} \\
Cutmix-Mixup switch prob & \multicolumn{4}{c}{0.5} \\
Random erasing prob & \multicolumn{4}{c}{0.25} \\
Label smoothing & \multicolumn{4}{c}{0.1} \\
Peak stochastic depth rate & 0.025 & 0.2 & 0.4 & 0.6 \\
Random erasing prob & \multicolumn{4}{c}{0.25} \\
EMA decay rate & \multicolumn{4}{c}{None} \\
\whline
\end{tabular}
\end{table}

\end{document}